\documentclass[review]{elsarticle}

\usepackage{amsmath,amssymb,amsfonts}
\usepackage{algorithmic}
\usepackage{graphicx}
\usepackage{textcomp}
\usepackage{subcaption}
\usepackage{multirow}
\usepackage{fancybox}
\usepackage{adjustbox}

\usepackage{lineno,hyperref}
\modulolinenumbers[1]

\journal{Journal of Biomedical Informatics}

\begin{document}

\begin{frontmatter}

\title{Multi-View Self-Attention for Interpretable Drug-Target Interaction Prediction}%\tnoteref{mytitlenote}}
%\tnotetext[mytitlenote]{}

\author[addruestc,addrsiping]{Brighter Agyemang\corref{mycorrespondingauthor}}
\cortext[mycorrespondingauthor]{Corresponding author}
\ead{brighteragyemang@gmail.com}
%\address{School of Computer Science and Engineering, University of Electronic Science and Technology of China, Chengdu, P. R. China}

\author[addruestc,addrsiping]{Wei-Ping Wu}
\author[addruestc,addrsiping]{Michael Yelpengne Kpiebaareh}
\author[addruestc,addrsiping]{Zhihua Lei}
\author[addruestc,addrsiping]{Ebenezer Nanor}
\author[addrsiping]{Lei Chen}


%\author[addruestc]{Global Customer Service\corref{mycorrespondingauthor}}

\address[addruestc]{School of Computer Science and Engineering, University of Electronic Science and Technology of China, Chengdu, P. R. China}
\address[addrsiping]{SipingSoft Co. Ltd., Tianfu Software Park, Chengdu, P. R. China}

\begin{abstract}
The drug discovery stage is a vital aspect of the drug development process and forms part of the initial stages of the development pipeline. In recent times,  machine learning-based methods are actively being used to model drug-target interactions for rational drug discovery due to the successful application of these methods in other domains.  In machine learning approaches, the numerical representation of molecules is critical to the performance of the model. While significant progress has been made in molecular representation engineering, this has resulted in several descriptors for both targets and compounds. Also, the interpretability of model predictions is a vital feature that could have several pharmacological applications. In this study, we propose a self-attention-based multi-view representation learning approach for modeling drug-target interactions.  We evaluated our approach using three benchmark kinase datasets and compared the proposed method to some baseline models. Our experimental results demonstrate the ability of our method to achieve competitive prediction performance and offer biologically plausible drug-target interaction interpretations.
\end{abstract}


\end{frontmatter}

%\linenumbers

\section{Introduction}\label{sec:introduction}
In the pharmaceutical sciences, drug discovery is the process of elucidating the roles of compounds in bioactivity for developing novel drugs. The drug discovery stage is vital to the drug development process and forms part of the initial stages of the development pipeline. In recent times, traditional \emph{in vivo} and \emph{in vitro} methods for analyzing bioactivities have been enhanced with automated methods such as large-scale High-Throughput Screening (HTS). The automation is motivated by the quest to reduce the cost and time-to-market challenges that are associated with the drug development process. The cost of developing a single drug is estimated to be 1.8 billion US dollars and could take 10-15 years to complete~\cite{Hopkins2009}. While HTS provides a better alternative to wet-lab experiments, it is time-consuming (takes about 2-3 years)~\cite{Lee2019} and requires advanced chemogenomic libraries. Also, with HTS, an exhaustive screening of the known human proteome and the $10^{60}$ synthetically feasible compounds is intractable~\cite{Rifaioglu2018, Polishchuk2013}. Additionally, HTS has a high failure rate~\cite{Doman2002}.

Lately, the availability of large-scale chemogenomic and pharmacological data (such as DrugBank~\cite{Knox2011}, KEGG~\cite{Kanehisa2012}, STITCH~\cite{Szklarczyk2016}, and ChemBL~\cite{Bento2014}, Davis~\cite{Davis2011}, KIBA~\cite{Tang2014}, PubChem~\cite{Kim2016}), coupled with advances in computational resources and algorithms have engendered the growth of the \emph{in silico} Virtual Screening (VS) domain. In silico methods have the potential to address the challenges mentioned above that plague HTS due to their ability to analyze assay data, unmask inherent relationships, and exploit such latent information for drug discovery tasks~\cite{Schierz2009}.

Consequently, there are several in silico proposals in the literature about DTI prediction. On account of data usage, structure-based methods, ligand-based approaches, and proteochemometric Modeling (PCM) constitute the taxonomy of existing in silico DTI studies. Structure-based methods use the 3D conformation of targets and compounds for bioactivity studies. Docking simulations are well-known instances of structure-based methods. Since the 3D conformation of several targets, such as G-Protein Coupled Receptors (GPCR) and Ion Channels (IC), are unknown, structure-based methods are limited in their application. They are also computationally expensive since a protein could assume multiple conformations depending on its rotatable bonds~\cite{Rifaioglu2018}. Ligand-based methods operate on the assumption that similar compounds would interact with similar targets and vice-versa. Hence, ligand-based methods perform poorly when a target (or compound) has a few known binding ligands (or targets) ($<100$).

On the other hand, PCM or chemogenomic methods, proposed in~\cite{LAPINSH2001180}, model interactions using a compound-target pair as input. Since PCM methods do not suffer from the drawbacks of ligand-based and structure-based methods, there have been several PCM-based DTI proposals in the literature~\cite{10.1093/bioinformatics/bti703,C4MD00216D,C5MB00088B}. Also, PCM methods can use a wide range of drug and target representations. Qiu et al. provide a well-documented growth of the PCM domain~\cite{Qiu2017}.

As regards computational methodologies, Chen et al. categorized existing models for DTI prediction into Network-based, Machine Learning (ML)-based, and other models~\cite{Chen2016}. Network-based methods approach the DTI prediction task using graph-theoretic algorithms where the nodes represent drugs and targets while the edges model the interactions between the nodes~\cite{Yamanishi2008}. As a corollary, the DTI prediction task becomes a link prediction problem. While network-based methods can work well even on datasets with few samples, they do not generalize to samples out of the training set, among other shortcomings. ML methods tackle the DTI prediction problem by training a parametric or non-parametric model iteratively with an independent and identically distributed training set made up of drug-target pairs using supervised, unsupervised, or semi-supervised algorithms. 
%Probabilistic Matrix Factorization (MF) of an interaction matrix and certain forms of similarity-based methods also exist in the domain~\cite{Cobanoglu2013, Chen2013}.

Similarity-based and feature-based methods are the main ML approaches in the literature. Similarity-based methods leverage the drug-drug, target-target, and drug-target similarities to predict new interactions~\cite{Shi2015, Perualila-Tan2016, Pahikkala2015}. Feature-based methods represent each drug or target using a numerical vector, which may reflect the entity's properties such as physicochemical features. The numerical vectors are then used to train an ML model to predict unknown interactions. 
%Sachdev et al. provide a thorough discussion of the feature-based DTI methods~\cite{Sachdev2019}. Additionally, some proposals combine feature-based and similarity-based methods to model interactions~\cite{Liu2015, He2017}. Due to the recent success of the Deep Learning (DL) domain, a form of ML, in areas such as computer vision~\cite{Xu2015} and Natural Language Processing (NLP)~\cite{Bahdanau2014}, recent feature-based approaches have mainly been DL algorithms~\cite{Wallach2015, Kearnes2016, Gomes2017, Altae-Tran2017, Lee2019, Shin2019}.

In feature-based methods, the construction of numerical vectors from the digital forms of drugs or targets is highly influential on model performance. This process is called featurization. The 2D structure of a compound can be represented using a line notation algorithm, such as the Simplified Molecular Input Line Entry System (SMILES)~\cite{Weininger1988}. Likewise, a target can be encoded using amino-acid sequencing. The compound and target features can then be computed using libraries such as RDKit~\cite{Landrum2006} and ProPy~\cite{Cao2013}, respectively. Due to the recent success of Deep Learning (DL) in areas such as computer vision~\cite{Xu2015} and Natural Language Processing (NLP)~\cite{Bahdanau2014}, recent feature-based approaches have mainly been DL algorithms. In this study, we focus on some challenges that are typically associated with DL-based DTI models.

Firstly, there exist several molecular representations or descriptors for both targets and compounds. These representations include both predefined descriptors~\cite{Todeschini2010, Rifaioglu2018, Mahmud2020} and end-to-end representation learning using backpropagation~\cite{Lee2019, Duvenaud2015, Wallach2015, Kearnes2016, Gomes2017, Wu2018, Tsubaki2019, Shin2019}. Since the choice of descriptors or features significantly affects model skill, there is an inexorable dilemma for researchers in feature selection~\cite{CERETOMASSAGUE201558, Kogej2006}. In some instances, molecular descriptors offer complementary behaviors~\cite{Sawada2014, Soufan2016, Mahmud2020} and their performance tend to be task related~\cite{DUAN2010157}. 
%Hence, the integration of these predefined descriptors is common and espoused by researchers to construct feature vectors~\cite{Baltrusaitis2019, Rifaioglu2018}. 
%Although these descriptors tend to provide domain-related information; their predefined nature means they are unable to establish a closer relationship between the input and output space concerning the task at hand.

Secondly, most of the existing DTI studies in the literature have formulated the DTI prediction task as a Binary Classification (BC) problem. However, the nature of bioactivity is continuous. Also, DTI depends on the concentration of the two query molecules and their intermolecular associations~\cite{Pahikkala2015}. Indeed, it is rare to have a ligand that binds to only one target~\cite{Rifaioglu2018}. While the binary classification approach provides a uniform approach to benchmark DTI proposals in the domain using the GPCR, IC, Enzymes (E), and Nuclear Receptor (NR) datasets of~\cite{Yamanishi2008}, treating DTI prediction as a binding affinity prediction problem leads to the construction of more realistic datasets~\cite{Ozturk2018, Tang2014}. 
%Accordingly, the Metz \cite{Metz2011}, KIBA \cite{Tang2014}, and Davis \cite{Davis2011} datasets serve as the benchmark datasets for regression-based DTI proposals and their output values are measured in dissociation constant $(K_d)$, KIBA metric~\cite{Tang2014}, and inhibition constant $(K_i)$, respectively.
 Another significant attribute of regression datasets is that they do not introduce class-imbalance problems seen with the BC datasets mentioned above. The BC-based algorithms typically address the class-imbalance problem using sampling techniques~\cite{Mahmud2020} or assume samples without reported interaction information to be non-interacting pairs.
We argue that predicting continuous values enable the entire spectrum of interaction to be well-captured in developing DTI prediction models.

Thirdly, since in silico DTI models are typically not replacements for in vitro and in vivo validations, interpreting DTI predictions could be vital to guiding domain experts. However, the application of multiple levels of non-linear transformation of the input means that DL models do not lend themselves easily to interpretation. In some studies, less powerful alternatives such as decision trees and $L_1$ regularization of linear models have been used to achieve interpretability of prediction results~\cite{Pliakos2019, Tabei2012}. Recent progress in pooling and attention-based techniques~\cite{Bahdanau2014, Santos2016, Vaswani2017} have also aided the ability to gain insights into DL-based prediction results~\cite{YingkaiGao2018, Shin2019}. We reckon that such attention-based mechanisms offer a route to provide biologically plausible insights into DL-based DTI prediction models while leveraging the strength of DL-models. 
%Also, since attention-based methods can learn rich molecular representations, it could facilitate accurate predictions in other domains such as ligand-catalyst-target reactions~\cite{Rifaioglu2018}.

To this end, our contributions as follows:
\begin{itemize}
	\item We propose a multi-view self-attention-based architecture for learning the representation of compounds and targets from different unimodal descriptor schemes (both predefined descriptors and end-to-end representations) for DTI prediction.
	\item Our usage of neural attention enables our proposed approach to lend itself to interpretation and discovery of biologically plausible insights in compound-target interactions across multiple views.
	\item We also experiment with existing baselines to show how these seemingly different compound and target featurization proposals in the literature could be aggregated to leverage their complementary relationships for modeling DTIs.
\end{itemize}

The rest of our study is organized as follows: Section~\ref{sec:related_work} discusses related work. We formalize the DTI problem addressed in this study in Section~\ref{sec:problem}. This is followed by a discussion of our proposed approach in Section~\ref{sec:jova}. Section~\ref{sec:experiments} presents the experiments we conducted to assess our proposed approach and we discuss the results in Section~\ref{sec:results_and_discussion}. Finally, we conclude our work in Section~\ref{sec:conclusion}.

\section{Related  Work}\label{sec:related_work}
%We review some of the notable works which relate to our study in this section.

One of the seminal works on integrating unimodal representations of drugs and compounds is~\cite{Yamanishi2010}. The authors integrated the chemical, genomic, and pharmacological spaces and used graph theory to model bioactivities. Shi et al~\cite{Shi2015} also augmented similarity information with non-structural features to perform DTI prediction using a network-based approach. Luo et al.~\cite{Luo2017} proposed a DTI  model that learns the contextual and topological properties of drug, disease, and target networks. Likewise, Wang et al.~\cite{Wang2018} proposed a random forest-based DTI prediction model that integrates features from drug, disease, target, and side-effect networks learned using GraRep~\cite{Cao2015}. These network-based DTI models are not scalable to large datasets and not applicable to samples outside the dataset.

In another vein, the work in~\cite{Pahikkala2015} proposed a Kronecker RLS (KronRLS) method that predicts binding affinity measured in $K_d$ and $K_i$. SimBoost was proposed in~\cite{He2017} as a Gradient Boosting Trees (GBT)-based DTI prediction model. While KronRLS is a linear model, SimBoost can learn non-linear properties for predicting real-valued binding affinities. While~\cite{He2017} used a feature-engineering step to select compound-target features for GBT training, the work in~\cite{Mahmud2020} integrates different representations of a target and uses a feature-selection algorithm to construct representations for GBT training.

Furthermore, several DL methods have been proposed to learn the features of compounds and targets for DTI prediction~\cite{Wu2018, Wallach2015, Gomes2017, Kearnes2016}, whereas others have proposed DL models that take predefined features as inputs. The work in~\cite{Wen2017} proposed a deep-belief network to model interactions using ECFP and Protein Sequence Composition (PSC) of compounds and targets, respectively.~\cite{Yang2018} also proposed a DTI model that uses generative modeling to oversample the minority class in order to address the class imbalance problem. In~\cite{Lee2019}, the sequence of a target is processed using a Convolutional Neural Network (CNN), whereas a compound is represented using its structural fingerprint. The compound and target feature vectors are concatenated and serve as input to a fully connected DL model. CNNs sacrifice the temporal structure in the target sequence to capture local residue information.

In contrast,~\cite{YingkaiGao2018} used a Recurrent Neural Network (RNN) and Molecular Graph Convolution (MGC) to learn the representations of targets and compounds, respectively. These representations are then processed by a siamese network to predict interactions. A limitation of the approach in~\cite{YingkaiGao2018} is that extending it to multi-task networks require training several siamese endpoints. Additionally, ~\cite{Ozturk2018} proposed a DL model that predicts binding affinities given compound and protein encoding that are learned using backpropagation.
The work in~\cite{Shin2019} also proposed a self-attention based DL model that predicts binding affinities. Using self-attention enables atom-atom relationships in a molecule to be adequately captured. 

Nonetheless, these studies do not leverage multimodal representations of compounds and targets for DTI prediction and bioactivity interpretation.

\section{Problem Formulation}\label{sec:problem}
We consider the problem of predicting a real-valued binding affinity $y_i$ between a given compound $c_i$ and target $t_i$, $i\in\mathbb{R}$. The compound $c_i$ takes the form of a SMILES~\cite{Weininger1988} string, whereas the target $t_i$ is encoded as an amino acid sequence. The SMILES string of $c_i$ is an encoding of a chemical graph structure $d_i=\{V_i, E_i\}$, where $V_i$ is the set of atoms constituting $c_i$ and $E_i$ is a set of undirected chemical bonds between these atoms. Therefore, each data point in the training set is a tuple $<c_i,t_i,y_i>$. In this study, we refer to the SMILES of a compound and the amino acid sequence of a target as the `raw' form of these entities, respectively. 

In order to use the compounds and targets in VS models, their respective raw forms have to be quantized to reflect their inherent physicochemical or structural properties. Accurately representing such properties is vital to reducing the generalization error of ML-based DTI models. Therefore, we address the problem of a DTI model that has the capacity to leverage multiple existing end-to-end and predefined representations in a rational manner that enables biologically plausible interpretations.

\section{Joint View Attention for DTI prediction}\label{sec:jova}
We propose a Joint View self-Attention (JoVA) approach to learn rich representations from multiple unimodal representations of compounds and targets for modeling bioactivity. Such a technique is significant when one considers that there exist several molecular representations, and that other novel methods are likely to be proposed, in the domain.

\begin{figure}
	\centering
	\includegraphics[scale=.25]{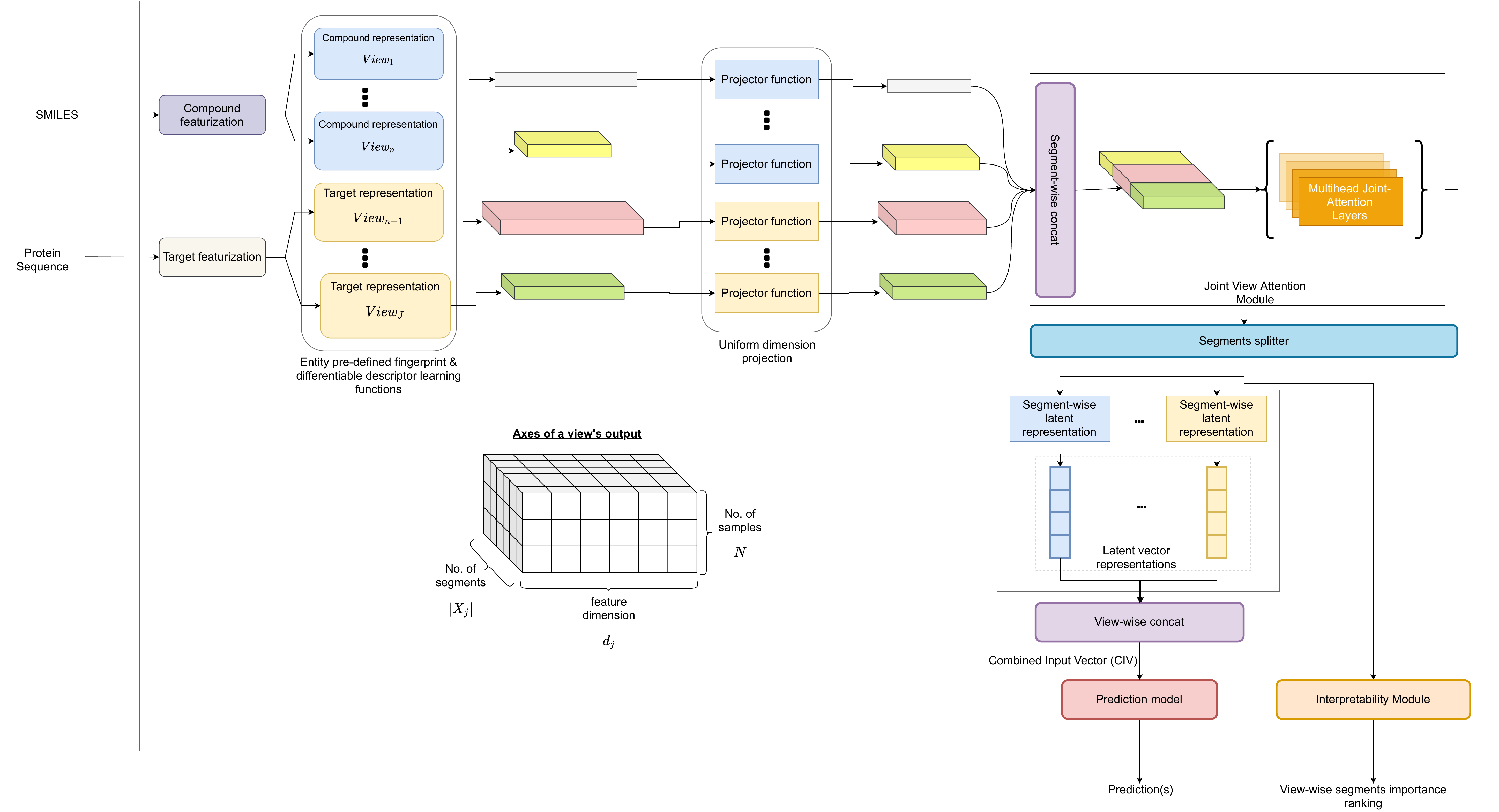}
	\caption{Joint View Attention(JoVA) for Drug-Target Interaction Prediction}
	\label{fig:jova}
\end{figure}

In Figure~\ref{fig:jova}, we present our proposed DL architecture for predicting binding affinities between compounds and targets. Before discussing the details of the architecture, we explain the terminology it uses:
\begin{itemize}
	\item Entity: this refers to a compound or target.
	\item View: this refers to a unimodal representation of an entity, such as ECFP for compounds or PSC for targets.
	\item Segment: for an entity represented as $X\in\mathbb{R}^{|X|\times d}$, we refer to the rows as the segments.
	\item Projector: projects an entity representation $X\in\mathbb{R}^{|X|\times d}$ into $X'\in\mathbb{R}^{|X|\times l}$, where $l\in\mathbb{R}$ is the latent space dimension.
	\item Concatenation function: We denote the concatenation (concat) function as $[\cdots]$.
	\item Combined Input Vector (CIV): a vector that is constructed by concatenating two or more vectors and used as the input of a function.
\end{itemize}

For a set of views $\mathcal{V}=\{v_1,v_2,...,v_J|J\in\mathbb{R}\}$, JoVA represents $v_j$ of an entity as $X_{v_j}\in\mathbb{R}^{|X_{v_j}|\times d_j}$ where $|X_{v_j}|$ denotes the number of elements that compose the entity and $d_j\in\mathbb{R}$ is the dimension of the feature vector of each of these elements of the $j$-th view. We write $X_{v_j}$ as $X_j$ in subsequent discussions to simplify notation. For a compound, the segments are the atoms, whereas a window of n-gram subsequences is a segment of a target. Note that in the case where the result of an entity featurization is a vector (e.g., ECFP and PSC), this is seen as $X_j\in\mathbb{R}^{1\times d_j}$. Thus, $|X_j|=1$.

Thereafter, a projection function $p_j$ of $v_j$ projects $X_j$ into a latent space of dimension $l$ to get $X_j'\in\mathbb{R}^{|X|\times l}$. Here, the dimension of each projection function is $l$. We refer to this operation as the latent dimension projection. We use the format (seg. denotes segment(s)),
\small{\begin{verbatim}
	(No. of seg., No. of samples, seg. dimension)
	\end{verbatim}}
to organize $N$ samples at this stage, employing zero-padding where necessary due to possible variation of the number of segments in a batch. This data structure follows the usual NLP tensor representation format where the number of segments is referred to as sequence length. Hence, the output of $v_j$ for a single entity is written as $X_j\in\mathbb{R}^{|X|\times 1\times l}$. This enables the concatenation of all projected representations to form the joint representation $\bar{X}=[X_1',X_2',...,X_J'], \bar{X}\in\mathbb{R}^{K\times 1\times l}$ where
\begin{equation}
K=\sum_{j=1}^J|X_j|.
\end{equation}
$\bar{X}$ then serves as the input to the joint view attention module. Since we use a single data point in our discussion, we write $\bar{X}\in\mathbb{R}^{K\times l}$ in subsequent discussions.

\begin{figure}[]
	\centering
	\includegraphics[scale=.6]{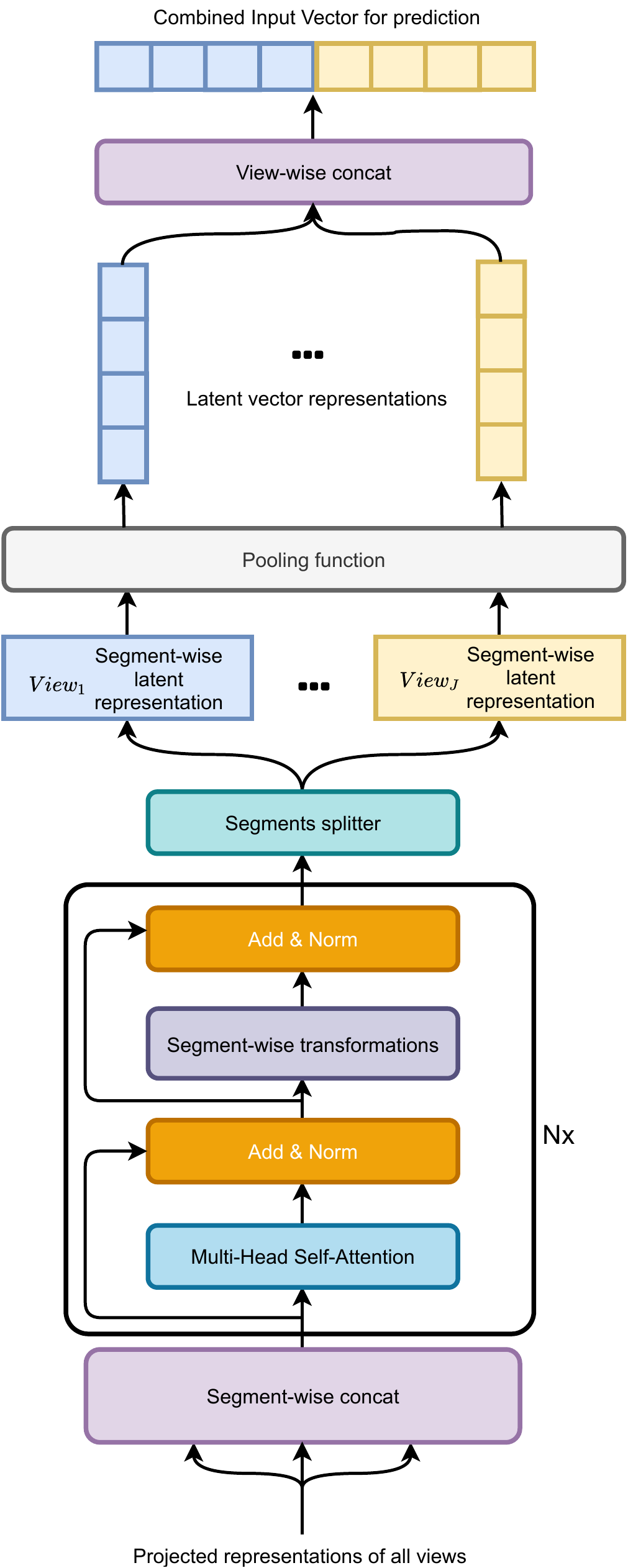}
	\caption{Architecture for constructing of the Combined Input Vector (CIV) using self-attention and pooling given the set of projected unimodal representations.}
	\label{fig:jova_attn}
\end{figure}

Figure~\ref{fig:jova_attn} illustrates the detailed processes between the segment-wise concat and view-wise concat layers of Figure~\ref{fig:jova}. Given the multi-view representation of an entity  $\bar{X}$, we apply a multihead self-attention mechanism and segment-wise input transformation~\cite{Vaswani2017}. An attention mechanism could be thought of as determining the relationships between a query and a set of key-value pairs to compute an output. Here, the query, keys, values, and outputs are vectors. Therefore, given a matrix of queries $Q$, a matrix of keys $K$, and a matrix of values $V$, the output of the attention function is expressed as,
\begin{equation}
Attention(Q,K,V)=softmax\left(\frac{QK^T}{\sqrt{d_k}}\right)V
\label{eq:dot_prod_attn}
\end{equation}
where $d_k$ is the dimension of $K$. In self-attention, we set $\bar{X}$ as $Q$, $K$, and $V$. The use of $\bar{X}$ as query, key, and value enables different unimodal segments to be related to all other views to compute the final representation of the compound-target pair. Intuitively, the individual graph of the views are merged to form a fully-connected graph where each node updates its representation by considering features of all other nodes using self-attention. Thus, each view becomes aware of all other views in learning its representation. This method therefore extends the two-way attention mechanism~\cite{YingkaiGao2018} to multiple unimodal representations. A single computation of equation~\ref{eq:dot_prod_attn} is referred to as a `head'.

In order to learn a rich representation of a compound-target pair, $\bar{X}$ is linearly projected into different subspaces, and the attention representation of each projection is computed after that. The resulting attention outputs are concatenated and also linearly projected to compute the output of the multihead sub-layer. For a set of self-attention heads $H=\{h_1,h_2,...,h_{|H|}\}$, the multihead function is expressed as,
\begin{equation}
Multihead(Q,K,V) = concat(H)W^O 
\end{equation}
where $h_i=Attention\left(QW_i^Q,KW_i^K,VW_i^V\right)$, $W_i^Q\in\mathbb{R}^{l\times d_k}$, $W_i^K\in\mathbb{R}^{l\times d_k}$, $W_i^V\mathbb{R}^{l\times d_v}$, $d_v$ is the dimension of $V$, and $W^O\in\mathbb{R}^{|H|d_v\times l}$.

Additionally, a segment-wise transformation sub-layer is used to transform each segment of the multihead attention sub-layer output non-linearly. Specifically, we compute
\begin{equation}
\hat{X} = ReLU(a_iW_1+b_1)W_2+b_2
\end{equation}
where $a_i$ denotes the $i$-th segment, $W_1\in\mathbb{R}^{l\times d_{seg}}$, $W_2\in\mathbb{R}^{d_{seg}\times l}$. We set $d_{seg}=2048$ in this study, same as found in~\cite{Vaswani2017}.

Furthermore, the Add and Norm layers in Figure~\ref{fig:jova_attn} implements a residual connection around the multihead and segment-wise transformation sublayers. This is expressed as $layerNorm(a_i+sublayer(a_i))$.


At the segments splitter layer, $\hat{X}$ is split into the constituting view representations $\{\hat{X}_1,\hat{X}_2,..., \hat{X}_J\}$. Note that $\hat{X}_j\in\mathbb{R}^{|X|\times l}$ for a single sample. The interpretability module then takes the updated representations of a view's segments $\hat{X}_j$ and rank these segments using the norm of their feature vectors to determine the influential segments of the view. 

To construct the final vector representation $\nu_j$ out of $\hat{X}_j$, pooling functions could then be applied to each view's representation. This enables our approach to be independent of the number of segments of each view, which could vary among samples. In this study, $\nu_j\in \mathbb{R}^l$ is computed as,
\begin{equation}
\nu_j = \sum_{i=1}^m\hat{X}_j^{(i)}
\end{equation}
where $m=|\hat{X}_j|$ and $\hat{X}_j^{(i)}$ denotes the $i$-th segment of $\hat{X}_j$. The view-wise concat layer subsequently computes the final representation of the compound-target pair as the concatenation of $[\nu_1,\nu_2,...,\nu_J]$ to get $\mathbf{x}\in\mathbb{R}^{Jl}$. We refer to $\mathbf{x}$ as the Combined Input Vector (CIV). The CIV therefore becomes the input to a prediction model. In our implementation of JoVA, the prediction model is a FCNN with 2-3 hidden layers.

\section{Experiments Design}\label{sec:experiments}
In this section, we present the experiments used to evaluate our proposed approach for DTI prediction. The compound and target featurization methods that we used in our experiments are briefly discussed in Section 1 of the supplementary document. We trained two variants of our proposed architecture. The first variant was trained using the ECFP8-GraphConv compound featurization methods and RNN-PSC target featurization methods. The second variant uses the ECFP8-GNN compound featurization methods and the PSC target featurization method. We denote the first and second variants as JoVA1 and JoVA2, respectively, in our discussions hereafter.
%Nonetheless, our proposed method is not restricted to this combination of compound and target featurization methods.

\subsection{Datasets}
\begin{figure}[!h]
	\centering	
	\subcaptionbox{\label{fig:davis_dist}}{
		\includegraphics[width=.6\linewidth]{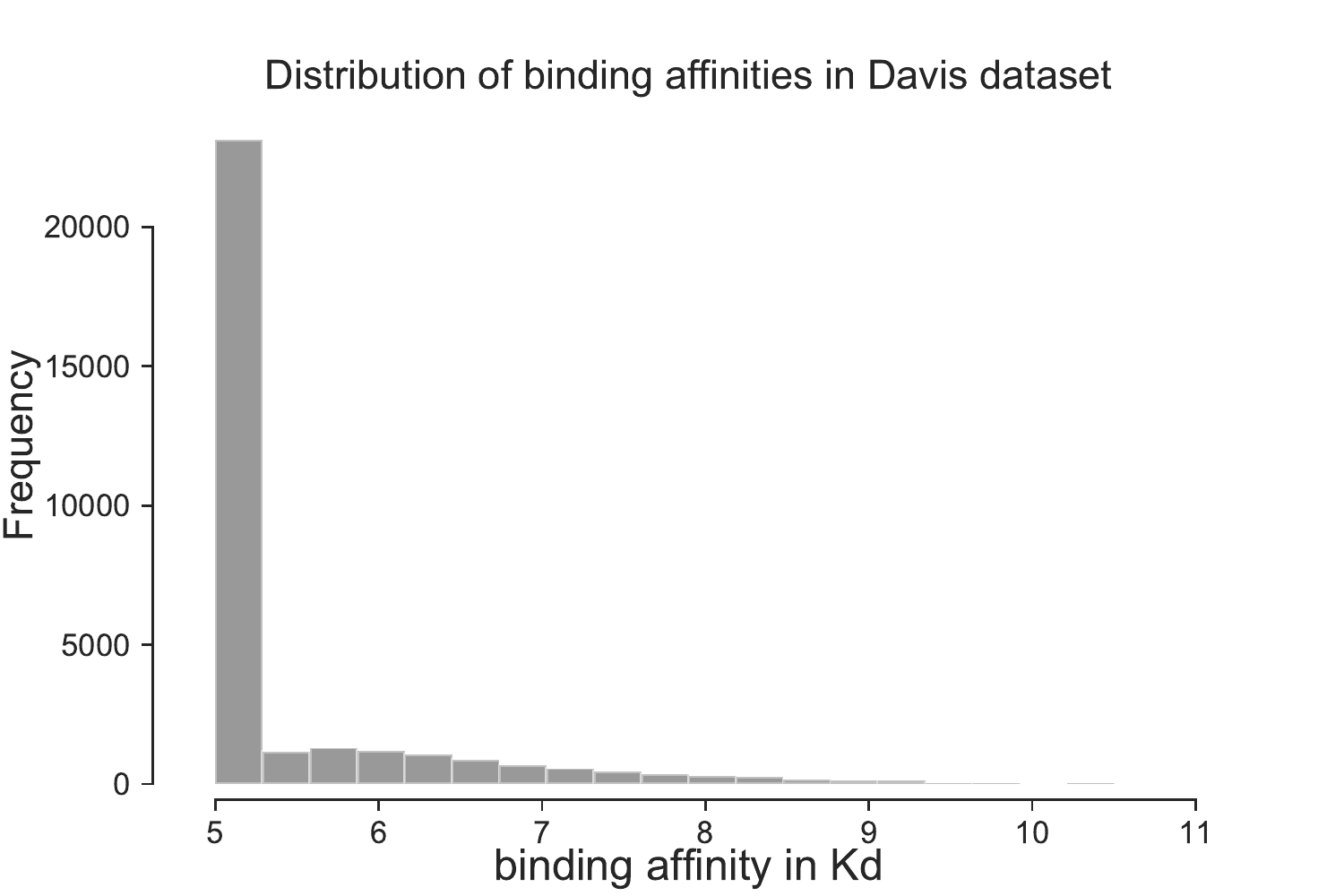}}\hspace{1em}%
	\subcaptionbox{\label{fig:metz_dist}}{\includegraphics[width=.6\linewidth]{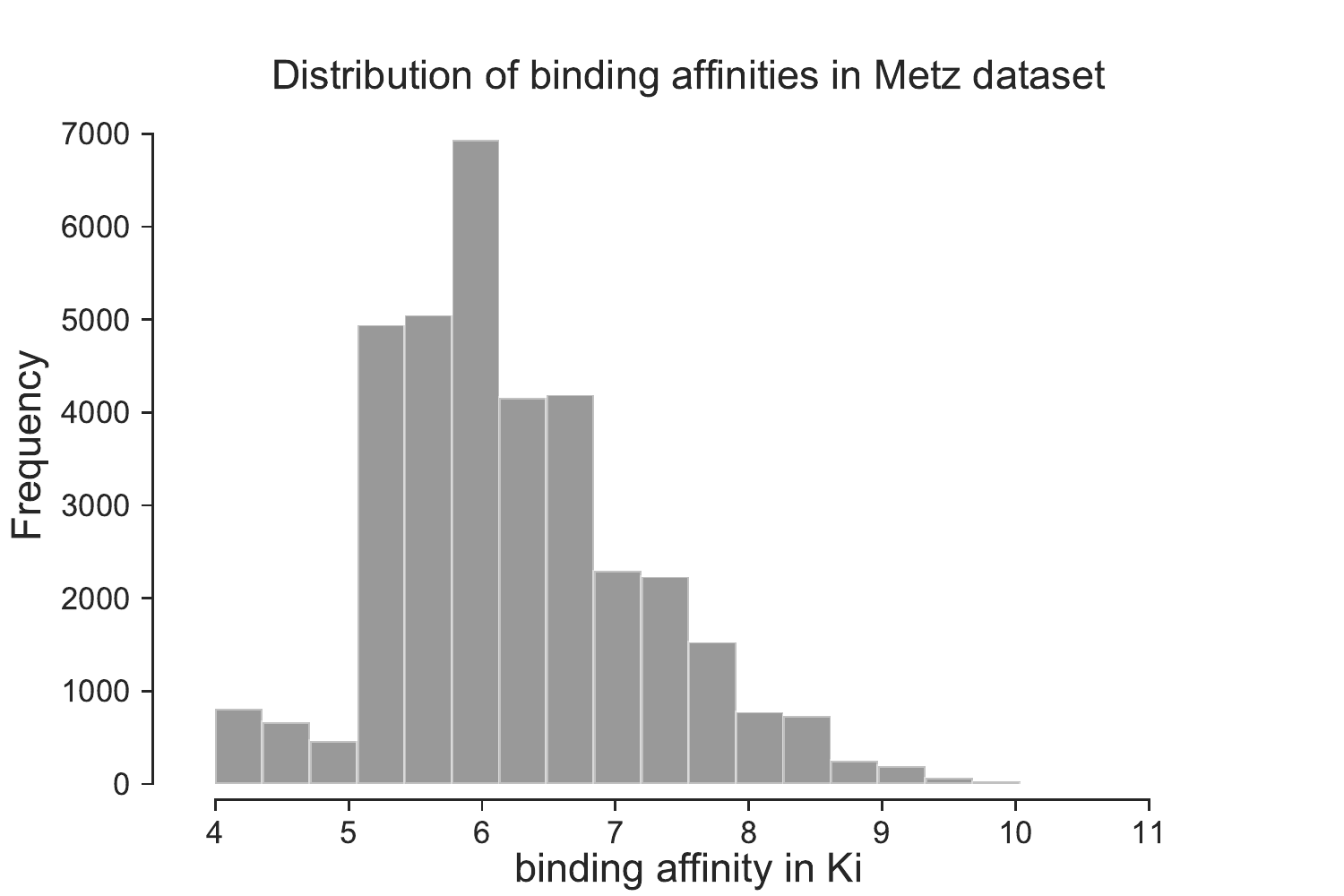}}
	\subcaptionbox{\label{fig:kiba_dist}}{\includegraphics[width=.6\linewidth]{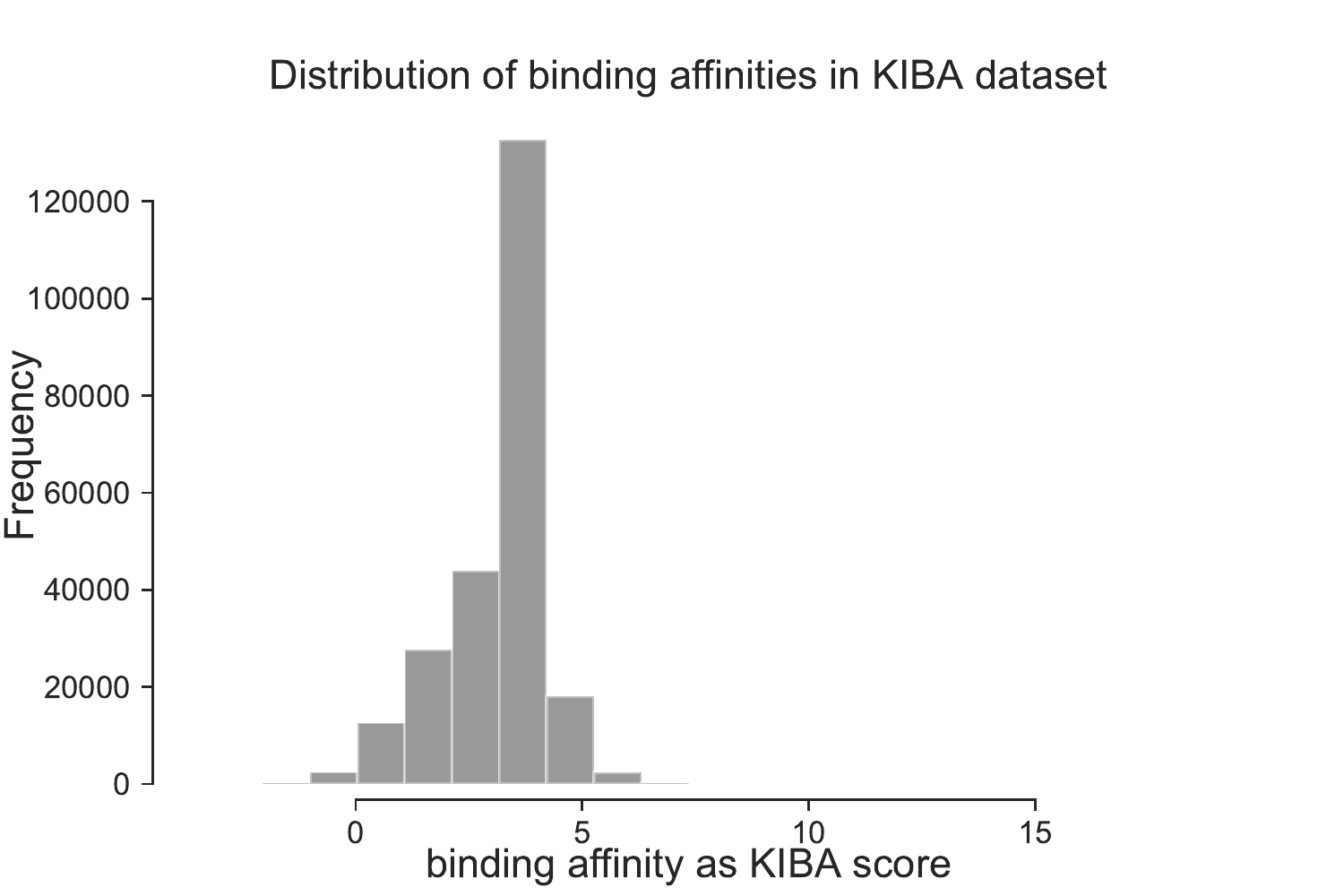}}\hspace{1em}%
	\caption{Distribution of the binding affinities (labels) in the Davis, Metz, and KIBA datasets used in our experiments.}
	\label{fig:data_dists}
\end{figure}

The benchmark datasets used in this study are the Metz \cite{Metz2011}, KIBA \cite{Tang2014}, and Davis \cite{Davis2011} datasets. Their output values are measured in dissociation constant $(K_d)$, KIBA metric~\cite{Tang2014}, and inhibition constant $(K_i)$, respectively. These are Kinase datasets that have been applied to benchmark previous DTI studies using the regression problem formulation~\cite{Pahikkala2015,  Ozturk2016, He2017, Feng2018, Shin2019}. Members of the Kinase family of proteins play active roles in cancer, cardiovascular, and other inflammatory diseases. However, their similarity makes it challenging to discriminate within the family. This similarity results in target promiscuity problems for binding ligands and, as a result, presents a challenging prediction task for ML models~\cite{Pahikkala2015}.  We used the version of these benchmark datasets curated by~\cite{Feng2018}. In~\cite{Feng2018}, a filter threshold is applied to each dataset for which compounds and targets with a total number of samples not above the threshold are removed. We maintain these thresholds in our study. The summary of these datasets, after filtering, is presented in table~\ref{tab:datasets}. Figure~\ref{fig:data_dists} shows the distribution of the binding affinities for the datasets. Although we used Kinase datasets to assess our proposed method, any DTI dataset that follows the problem formulation in Section~\ref{sec:problem} is applicable.

%\begin{figure}[h]
%	\centering	
%	\subcaptionbox{\label{fig:davis_dist}}{\includegraphics[width=.6\linewidth]{davis.pdf}}\hspace{1em}%
%	\subcaptionbox{\label{fig:metz_dist}}{\includegraphics[width=.6\linewidth]{metz.pdf}}
%	\subcaptionbox{\label{fig:kiba_dist}}{\includegraphics[width=.6\linewidth]{kiba.pdf}}\hspace{1em}%
%	\caption{Distribution of the binding affinities (labels) in the Davis, Metz, and KIBA datasets used in our experiments.}
%	\label{fig:data_dists}
%\end{figure}

\begin{table}[]
	\caption{Dataset sizes}
	\label{tab:datasets}
	\centering
	\begin{tabular}{|l|l|l|l|l|}
		\hline
		\textbf{Dataset} & \textbf{\begin{tabular}[c]{@{}l@{}}Number of \\ compounds\\/drugs\end{tabular}} & \textbf{\begin{tabular}[c]{@{}l@{}}Number \\ of targets\end{tabular}} & \textbf{\begin{tabular}[c]{@{}l@{}}Total number\\ of pair samples\end{tabular}} & \textbf{\begin{tabular}[c]{@{}l@{}}Filter \\ threshold \\ used\end{tabular}} \\ \hline
		Davis            & 72                                                                            & 442                                                                   & 31824                                                                           & 6                                                                    \\ \hline
		Metz             & 1423                                                                          & 170                                                                   & 35259                                                                           & 1                                                                    \\ \hline
		KIBA             & 3807                                                                          & 408                                                                   & 160296                                                                          & 6                                                                    \\ \hline
	\end{tabular}
\end{table}

\begin{table}[]
	\centering
	\caption{Simulation hardware specifications}
	\label{tab:hdw_specs}
	\begin{tabular}{|l|l|l|l|l|}
		\hline
		\multicolumn{1}{|c|}{\textbf{Model}}                               & \multicolumn{1}{c|}{\textbf{\# Cores}} & \multicolumn{1}{c|}{\textbf{\begin{tabular}[c]{@{}c@{}}RAM \\ (GB)\end{tabular}}} & \multicolumn{1}{c|}{\textbf{\begin{tabular}[c]{@{}c@{}}Avail. \\ GPUs\end{tabular}}} & \multicolumn{1}{c|}{\textbf{\begin{tabular}[c]{@{}c@{}}\# GPUs \\ used\end{tabular}}} \\ \hline
		\begin{tabular}[c]{@{}l@{}}Intel Xeon \\ CPU E5-2687W\end{tabular} & 48                                     & 128                                                                             & \begin{tabular}[c]{@{}l@{}}1 GeForce\\  GTX 1080\end{tabular}                        & 1                                                                                     \\ \hline
		\begin{tabular}[c]{@{}l@{}}Intel Xeon \\ CPU E5-2687W\end{tabular} & 24                                     & 128                                                                             & \begin{tabular}[c]{@{}l@{}}4 GeForce \\ GTX 1080Ti\end{tabular}                      & 2                                                                                     \\ \hline
	\end{tabular}
\end{table}

\subsection{Baselines}
We compare our proposed approach to the works in~\cite{Pahikkala2015, He2017, Feng2018, Tsubaki2019} and our previously proposed IVPGAN approach in~\cite{Agyemang2019}. We provide a review of each of the baseline methods\footnote{\url{https://github.com/bbrighttaer/jova_baselines}} in Section 2 of the accompanying supplementary document. Since the method in~\cite{Tsubaki2019} was trained for binary classification, we replaced the endpoint of the model in~\cite{Tsubaki2019} with a regression layer in our experiments. The labels we give to~\cite{Pahikkala2015, He2017} and~\cite{Tsubaki2019} are KronRLS, SimBoost, and CPI-Reg, respectively. \cite{Feng2018} proposed two models for DTI prediction: PADME-ECFP and PADME-GraphConv. The PADME-ECFP model integrates ECFP4 of a compound and PSC of a target to construct the feature vector of a query compound and target. In our experiments, we used ECFP8 instead and refer to this variant of PADME-ECFP as ECFP8-PSC. The justification for this modification is discussed in~\cite{Agyemang2019}. Likewise, the PADME-GraphConv model is referred to as GraphConv-PSC in our study since it uses molecular graph convolution to learn the representation of a compound and PSC for a target feature vector. SimBoost and KronRLS were implemented as XGBoost and Numpy models, respectively.  

\subsection{Model Training and Evaluation}
In our experiments, we used a 5-fold Cross-Validation (CV) model training approach. Each CV fold was divided into train, validation, and test sets. The validation and test sets were used for hyperparameter search and model evaluation, respectively. The following three splitting schemes were used:
\begin{itemize}
	\item \textbf{Warm split}: Every drug or target in the validation and test sets is encountered in the training set.
	\item \textbf{Cold-drug split}: Every compound in the validation and test sets is absent from the training set.
	\item \textbf{Cold-target split}: Every target in the validation and test set is absent from the training set.
\end{itemize}
Since \emph{cold-start} predictions are typically found in DTI use cases, the cold splits provide realistic and more challenging evaluation schemes for the models. 

We used Soek\footnote{\url{https://github.com/bbrighttaer/soek}}, a Python library based on Scikit-Optimize, to determine the best performing hyperparameters for each model on the validation set. We used the warm split scheme of the Davis dataset for the hyperparameter search. The best hyperparameters were then kept fixed for all other split schemes across the Metz and KIBA datasets. This was done due to the enormous time and resource requirements needed to repeat the search in each case of the experiment. 

Nonetheless, in the case of each SimBoost experiment, we searched for the best performing latent dimension of the Matrix Factorization stage for each dataset. This exception was necessary because we realized that the best MF latent dimension found on the Davis dataset produced poor MF on the other datasets which heavily influenced SimBoost performance. Thus, this exception made SimBoost better positioned to perform better than the other models on the warm split schemes of the Metz and KIBA datasets.

As regards evaluation metrics, we measured the Root Mean Squared Error (RMSE) and Pearson correlation coefficient ($ R^2 $) on the test set in each CV-fold. Additionally, we measured the Concordance Index (CI) on the test set, as proposed by~\cite{Pahikkala2015}. 

We followed the averaging CV approach, where the reported metrics are the averages across the different folds. We also repeated the CV evaluation for different random seeds to minimize the effect of randomness in the reported results. After that, all metrics were averaged across the random seeds.

\begin{table*}[]
	\centering
	\caption{Performance of regression on benchmark datasets measured in RMSE (smaller is better). The best scores are marked in bold and the standard deviation values are in parenthesis.}
	\label{tab:rmse}
	\resizebox{\linewidth}{!}{
		\begin{tabular}{|l|l|l|l|l|l|l|l|l|l|}
			\hline
			\multicolumn{10}{|c|}{RMSE}                                                                                                                                                                                                        \\ \hline
			\textbf{Dataset}       & \textbf{CV Split Type} & \textbf{ECFP8-PSC}     & \textbf{GraphConv-PSC} & \textbf{KronRLS} & \textbf{SimBoost}      & \textbf{CPI-Reg} & \textbf{IVPGAN}        & \textbf{JoVA1}         & \textbf{JoVA2}         \\ \hline
			\multirow{3}{*}{Davis} & Warm                   & \textbf{0.206 (0.084)} & 0.268 (0.0819)         & 0.614 (0.016)    & 0.218 (0.113)          & 0.756 (0.196)    & 0.206 (0.085)          & 0.260 (0.084)          & 0.260 (0.063)          \\ \cline{2-10} 
			& Cold Drug              & 0.351 (0.169)          & 0.383 (0.165)          & 0.766 (0.090)    & -                      & 0.763 (0.216)    & 0.241 (0.129)          & \textbf{0.318 (0.112)} & 0.323 (0.107)          \\ \cline{2-10} 
			& Cold Target            & 0.265 (0.133)          & 0.328 (0.118)          & 0.564 (0.032)    & -                      & 0.752 (0.215)    & \textbf{0.245 (0.119)} & 0.290 (0.127)          & 0.307(0.108)           \\ \hline
			\multirow{3}{*}{Metz}  & Warm                   & 0.343 (0.058)          & 0.401 (0.043)          & 0.796 (0.008)    & \textbf{0.234 (0.102)} & 0.738 (0.114)    & 0.295 (0.076)          & 0.345 (0.041)          & 0.302 (0.041)          \\ \cline{2-10} 
			& Cold Drug              & 0.348 (0.139)          & 0.440 (0.095)          & 0.786 (0.032)    & -                      & 0.753 (0.129)    & \textbf{0.312 (0.114)} & 0.377 (0.092)          & 0.340 (0.098)          \\ \cline{2-10} 
			& Cold Target            & 0.622 (0.175)          & 0.525 (0.067)          & 0.822 (0.034)    & -                      & 0.730 (0.123)    & 0.551 (0.102)          & 0.366 (0.072)          & \textbf{0.356 (0.068)} \\ \hline
			\multirow{3}{*}{KIBA}  & Warm                   & 0.366 (0.097)          & 0.488 (0.124)          & 0.671 (0.008)    & \textbf{0.346 (0.064)} & 0.646 (0.125)    & 0.414 (0.083)          & 0.388 (0.081)          & 0.380 (0.105)          \\ \cline{2-10} 
			& Cold Drug              & 0.418 (0.154)          & 0.500 (0.158)          & 0.673 (0.025)    & -                      & 0.670 (0.205)    & 0.427 (0.113)          & \textbf{0.401 (0.140)} & 0.418 (0.186)          \\ \cline{2-10} 
			& Cold Target            & 0.460 (0.124)          & 0.501 (0.121)          & 0.712 (0.027)    & -                      & 0.648 (0.128)    & 0.483 (0.093)          & 0.411 (0.101)          & \textbf{0.397 (0.097)} \\ \hline
		\end{tabular}
	}
\end{table*}

\begin{table*}[]
	\centering
	\caption{Performance of regression on benchmark datasets measured in CI (larger is better). The best scores are marked in bold and the standard deviation values are in parenthesis.}
	\label{tab:ci}
	\resizebox{\linewidth}{!}{
		\begin{tabular}{|l|l|l|l|l|l|l|l|l|l|}
			\hline
			\multicolumn{10}{|c|}{\textbf{Concordance Index}}                                                                                                                                                                                   \\ \hline
			\textbf{Dataset}       & \textbf{CV Split Type} & \textbf{ECFP8-PSC}     & \textbf{GraphConv-PSC} & \textbf{KronRLS} & \textbf{SimBoost}      & \textbf{CPI-Reg} & \textbf{IVPGAN}        & \textbf{JoVA1} & \textbf{JoVA2}         \\ \hline
			\multirow{3}{*}{Davis} & Warm                   & 0.965 (0.022)          & 0.946 (0.025)          & 0.890 (0.003)    & 0.968 (0.026)          & 0.762 (0.049)    & \textbf{0.970 (0.020)} & 0.951 (0.020)  & 0.948 (0.019)          \\ \cline{2-10} 
			& Cold Drug              & 0.929 (0.059)          & 0.909 (0.062)          & 0.732 (0.037)    & -                      & 0.720 (0.064)    & \textbf{0.952 (0.050)} & 0.924 (0.034)  & 0.920 (0.033)          \\ \cline{2-10} 
			& Cold Target            & 0.946 (0.041)          & 0.930 (0.038)          & 0.869 (0.008)    & -                      & 0.760 (0.057)    & \textbf{0.958 (0.036)} & 0.934 (0.040)  & 0.936 (0.033)          \\ \hline
			\multirow{3}{*}{Metz}  & Warm                   & 0.898 (0.023)          & 0.859 (0.022)          & 0.772 (0.003)    & \textbf{0.943 (0.036)} & 0.687 (0.056)    & 0.912 (0.031)          & 0.883 (0.019)  & 0.908 (0.015)          \\ \cline{2-10} 
			& Cold Drug              & 0.890 (0.058)          & 0.839 (0.045)          & 0.726 (0.010)    & -                      & 0.647 (0.072)    & \textbf{0.906 (0.048)} & 0.870 (0.039)  & 0.889 (0.037)          \\ \cline{2-10} 
			& Cold Target            & 0.808 (0.048)          & 0.805 (0.036)          & 0.732 (0.015)    & -                      & 0.688 (0.057)    & 0.813 (0.040)          & 0.877 (0.029)  & \textbf{0.890 (0.026)} \\ \hline
			\multirow{3}{*}{KIBA}  & Warm                   & 0.864 (0.031)          & 0.814 (0.037)          & 0.799 (0.002)    & \textbf{0.882 (0.024)} & 0.692 (0.066)    & 0.836 (0.023)          & 0.852 (0.034)  & 0.860 (0.038)          \\ \cline{2-10} 
			& Cold Drug              & \textbf{0.837 (0.065)} & 0.795 (0.065)          & 0.723 (0.006)    & -                      & 0.627 (0.100)    & 0.829 (0.044)          & 0.836 (0.061)  & 0.833 (0.064)          \\ \cline{2-10} 
			& Cold Target            & 0.818 (0.050)          & 0.794 (0.047)          & 0.757 (0.009)    & -                      & 0.692 (0.068)    & 0.805 (0.042)          & 0.839 (0.043)  & \textbf{0.845 (0.042)} \\ \hline
		\end{tabular}
	}
\end{table*}

\begin{table*}[]
	\centering
	\caption{Performance of regression on benchmark datasets measured in $R^2$ (larger is better). The best scores are marked in bold and the standard deviation values are in parenthesis.}
	\label{tab:r2}
	\resizebox{\linewidth}{!}{
		\begin{tabular}{|l|l|l|l|l|l|l|l|l|l|}
			\hline
			\multicolumn{10}{|c|}{\textbf{$R^2$}}                                                                                                                                                                                              \\ \hline
			\textbf{Dataset}       & \textbf{CV Split Type} & \textbf{ECFP8-PSC} & \textbf{GraphConv-PSC} & \textbf{KronRLS} & \textbf{SimBoost}      & \textbf{CPI-Reg} & \textbf{IVPGAN}        & \textbf{JoVA1} & \textbf{JoVA2}         \\ \hline
			\multirow{3}{*}{Davis} & Warm                   & 0.933 (0.064)      & 0.894 (0.064)          & 0.639 (0.013)    & 0.928 (0.071)          & 0.259 (0.117)    & \textbf{0.935 (0.058)} & 0.906 (0.045)  & 0.903 (0.045)          \\ \cline{2-10} 
			& Cold Drug              & 0.814 (0.158)      & 0.757 (0.194)          & 0.264 (0.101)    & -                      & 0.200 (0.131)    & \textbf{0.879 (0.172)} & 0.833 (0.117)  & 0.827 (0.112)          \\ \cline{2-10} 
			& Cold Target            & 0.881 (0.126)      & 0.840 (0.124)          & 0.602 (0.031)    & -                      & 0.260 (0.133)    & \textbf{0.903 (0.111)} & 0.860 (0.133)  & 0.855 (0.106)          \\ \hline
			\multirow{3}{*}{Metz}  & Warm                   & 0.859 (0.056)      & 0.799 (0.054)          & 0.537 (0.012)    & 0.931 (0.065)          & 0.334 (0.155)    & \textbf{0.895 (0.062)} & 0.852 (0.045)  & 0.889 (0.034)          \\ \cline{2-10} 
			& Cold Drug              & 0.827 (0.152)      & 0.738 (0.126)          & 0.410 (0.027)    & -                      & 0.259 (0.183)    & \textbf{0.867 (0.115)} & 0.814 (0.105)  & 0.842 (0.101)          \\ \cline{2-10} 
			& Cold Target            & 0.600 (0.160)      & 0.653 (0.104)          & 0.422 (0.046)    & -                      & 0.337 (0.158)    & 0.657 (0.122)          & 0.826 (0.076)  & \textbf{0.849 (0.066)} \\ \hline
			\multirow{3}{*}{KIBA}  & Warm                   & 0.790 (0.085)      & 0.639 (0.117)          & 0.571 (0.012)    & \textbf{0.840 (0.073)} & 0.353 (0.166)    & 0.754 (0.062)          & 0.769 (0.085)  & 0.785 (0.089)          \\ \cline{2-10} 
			& Cold Drug              & 0.685 (0.180)      & 0.569 (0.189)          & 0.437 (0.028)    & -                      & 0.254 (0.209)    & \textbf{0.721 (0.120)} & 0.713 (0.167)  & 0.695 (0.168)          \\ \cline{2-10} 
			& Cold Target            & 0.673 (0.135)      & 0.612 (0.131)          & 0.418 (0.038)    & -                      & 0.354 (0.168)    & 0.661 (0.100)          & 0.746 (0.105)  & \textbf{0.755 (0.095)} \\ \hline
		\end{tabular}
	}
\end{table*}

\begin{figure}
	\centering
	\resizebox{\linewidth}{!}{
		\subcaptionbox{\label{fig:davis_qualitative}}
		{
			\includegraphics[scale=.4]{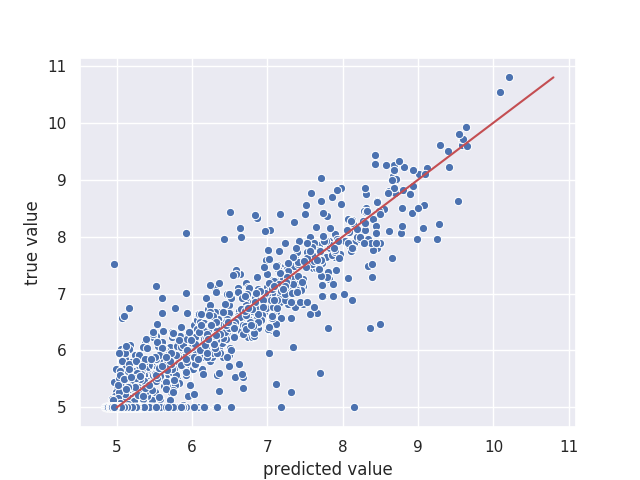}
			\includegraphics[scale=.4]{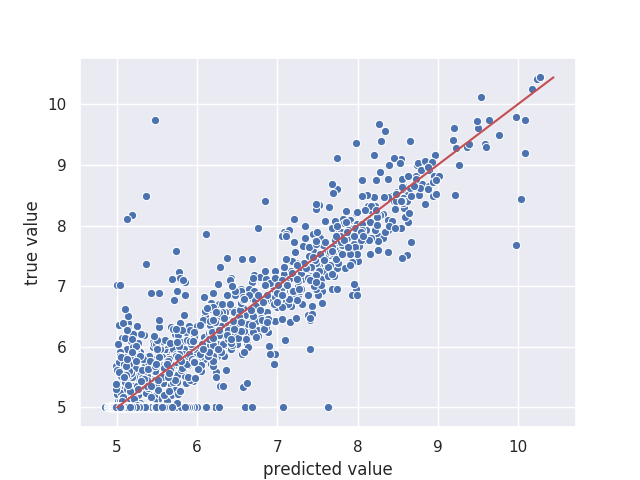}
			\includegraphics[scale=.4]{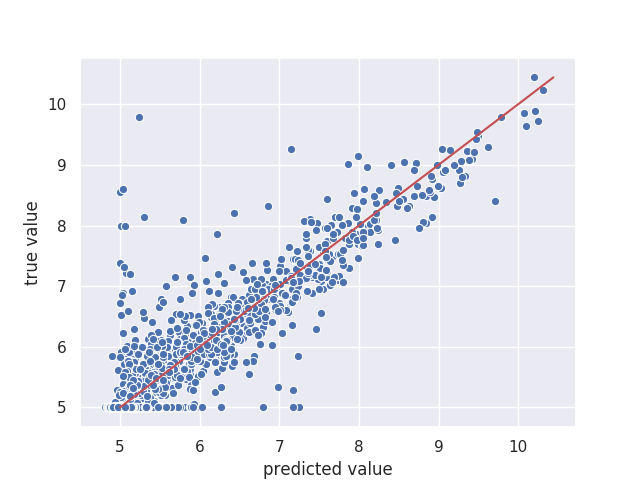}
		}
	}\hspace{1em}%
	\\
	\resizebox{\linewidth}{!}{
		\subcaptionbox{\label{fig:metz_qualitative}}
		{
			\includegraphics[scale=.4]{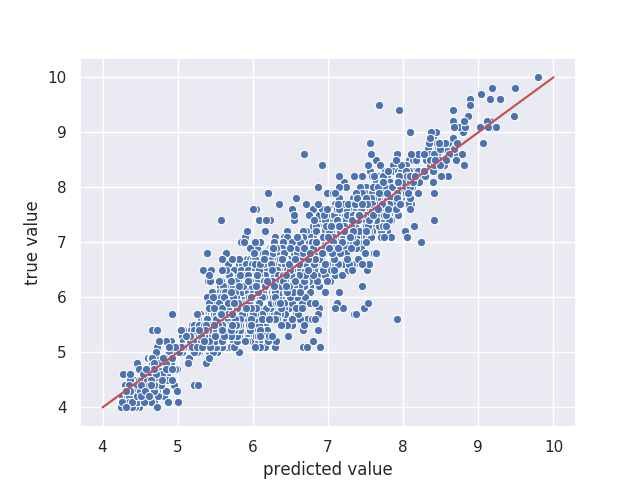}
			\includegraphics[scale=.4]{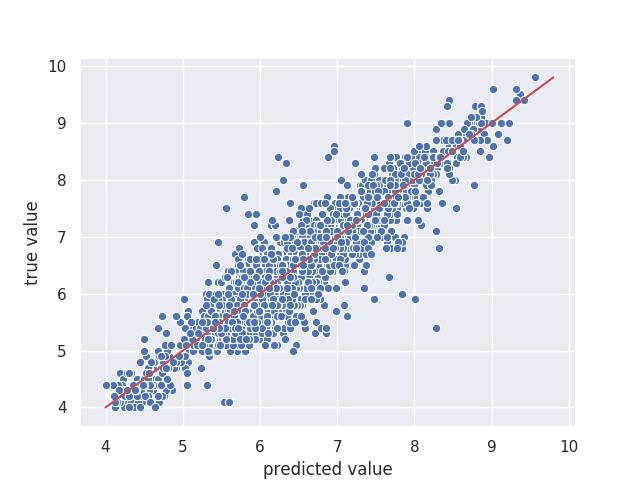}
			\includegraphics[scale=.4]{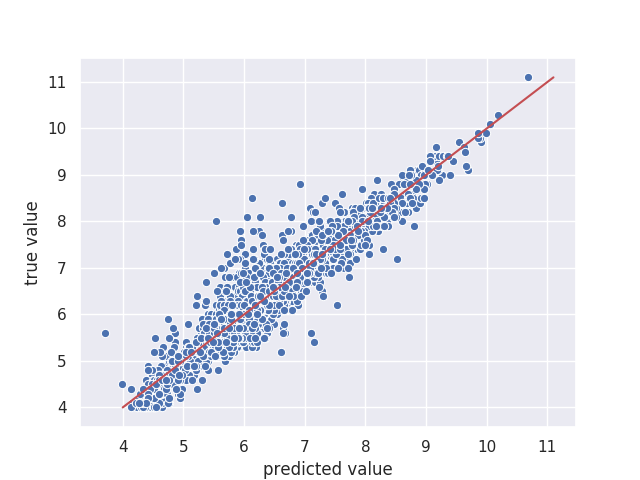}
		}
	}\hspace{1em}%
	\\
	\resizebox{\linewidth}{!}{
		\subcaptionbox{\label{fig:kiba_qualitative}}
		{
			\includegraphics[scale=.4]{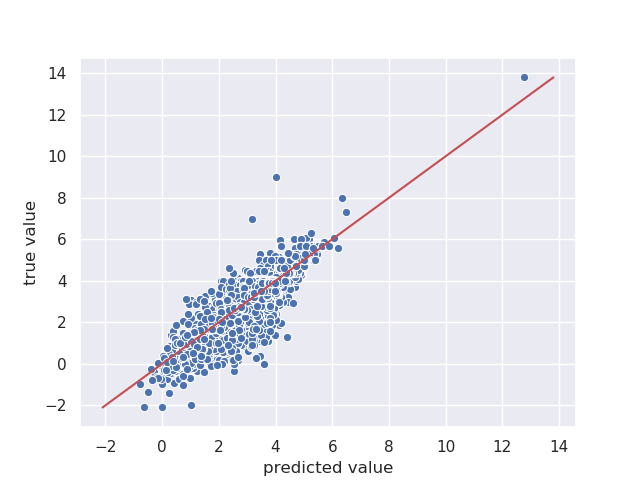}
			\includegraphics[scale=.4]{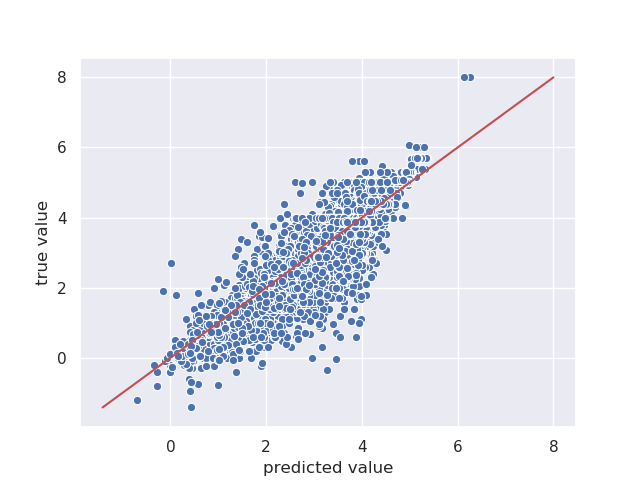}
			\includegraphics[scale=.4]{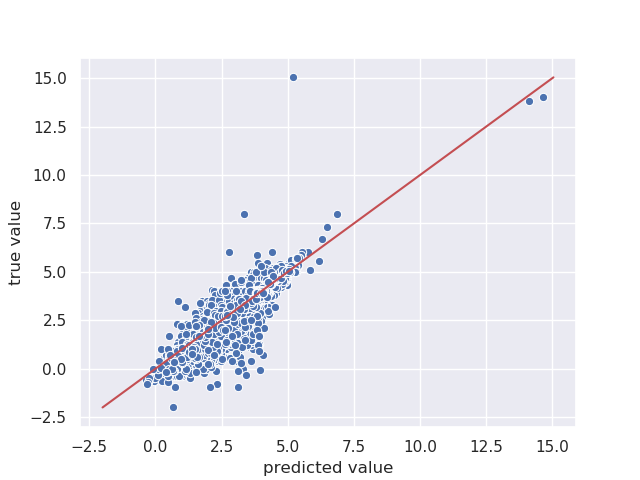}
		}
	}\hspace{1em}%
	
	\caption{\textbf{Predictions of JoVA1 on the benchmark dataset.} (From left to right) Columns 1, 2, and 3 show the warm split, cold-drug split, and cold-target split results, respectively. The first, second, and third rows correspond to the Davis, Metz, and KIBA datasets, respectively.}
	\label{fig:scatter_plots}
\end{figure}

\section{Results and Discussion}\label{sec:results_and_discussion}
In this section, we first discuss the performance of all baseline and JoVA models. Here, performance refers to the CI, RMSE, and $R^2$ scores of a model. The evaluation results of the models are presented in Tables~\ref{tab:rmse},\ref{tab:ci}, and \ref{tab:r2}. Smaller RMSE values connote better performance whereas larger CI and $R^2$ values indicate better performance.
We then discuss our findings on applying our proposed approach to case studies on predicting novel DTIs and interpretability.

\subsection{Prediction Performance}
On the whole, models that used multi-view representations attained the best performance in most of our experiments while the ECFP-PSC and SimBoost models achieved best performance in some few cases. This indicates the significance of integrating unimodal representations of compounds and targets to leverage the resulting complementary properties for DTI prediction. Although IVPGAN performed better than the JoVA models, especially on the CI and $R^2$ metrics, the JoVA models scored competitively to the best model and also offer the appealing ability to interpret model predictions by mapping attention outputs to the atoms of a compound or protein sequence segments.

Additionally, we realized that the cold split schemes that had fewer entities turned out to be the most challenging for the models. An instance is the cold drug split scheme on the Davis dataset. This phenomenon was more visible across the baseline models while the JoVA models usually exhibited a relatively stable performance on the three datasets. We reckon that the relative stability of the JoVA models' performance is due the proposed self-attention-based multi-view representation learning approach for constructing the feature vector of a query compound and target pair since the complementary relationships among the featurization methods are harnessed.

Despite CPI-Reg using an end-to-end representation learning for both compound and target, it performed worst than almost all the other models. We suggest this counter-intuitiveness indicates that purely end-to-end unimodal representation learning methods with big embedding matrices require large datasets to effectively learn good representations. Thus, the GraphConv-PSC model was less susceptible to this challenge since it used end-to-end representation learning for only compounds. Therefore, simpler models (in terms of the number of trainable parameters), could perform well on small datasets using predefined featurization methods such as ECFP and PSC.

As regards the traditional ML models, KronRLS recorded modest results for a linear model, whereas SimBoost achieved best results in some cases. We think that searching for the best MF latent dimension for each dataset contributed to the improvement in the results of SimBoost since \cite{He2017} shows the significance of the MF features to SimBoost predictions. It is also noteworthy that SimBoost's feature engineering phase renders it inapplicable to the cold splitting schemes.

In Figure~\ref{fig:scatter_plots} we show the scatter plots of the JoVA1 model's predictions on the three benchmark datasets used in this study across the split schemes. While the predictions on the Davis and Metz datasets had high variance, the predictions on the KIBA dataset are clustered in certain regions, similar to the data distributions in Figure~\ref{fig:data_dists}. This common trend in the plots of the predictions by the JoVA model also helps to qualitatively assess the stability of our proposed approach across the split schemes. The scatter plots of the other baseline models have been provided in Section 3 of the accompanying supplementary document.

Taken together, we think that using self-attention to align multiple unimodal representations of atoms and amino acid residues to each other provides a better representational capacity. Also, unlike the baseline models, JoVA provides a rational approach to obtain model interpretations by examining the outputs of the joint-view attention layer, as we demonstrate in Section~\ref{sec:interpretability}.

\subsection{DrugBank Case Study}
\begin{figure}
	\centering
	\cornersize{.1}
	%	\ovalbox{
	
	\subcaptionbox{\label{fig:brigatinib}}
	{
		\includegraphics[scale=.2]{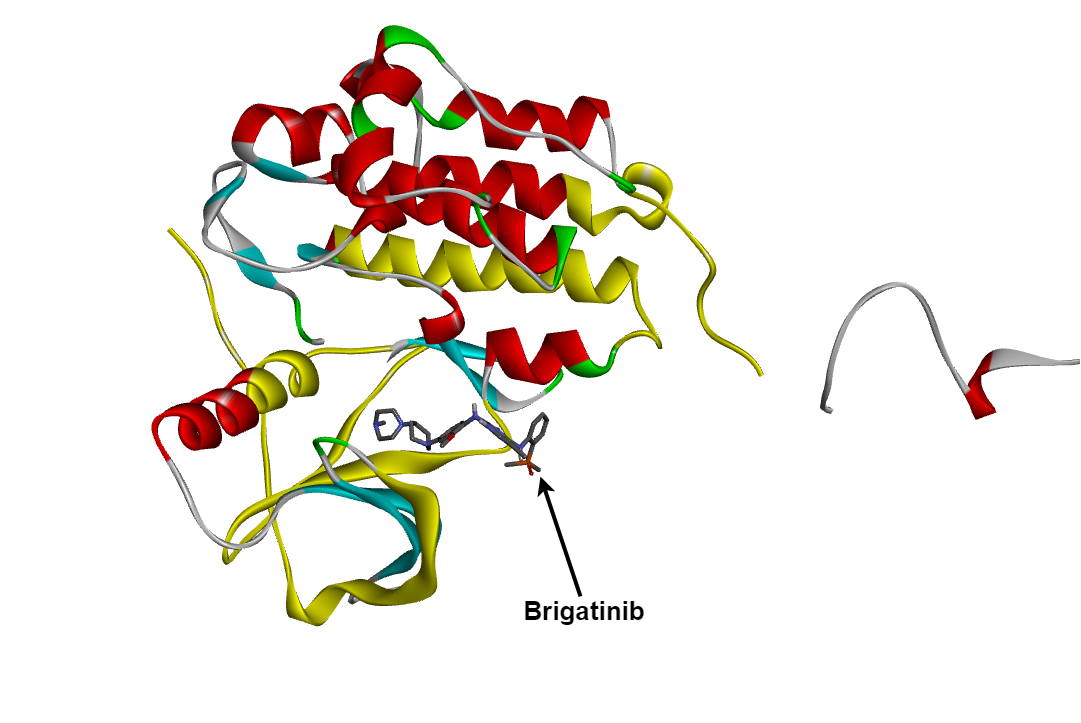}
		\includegraphics[scale=.2]{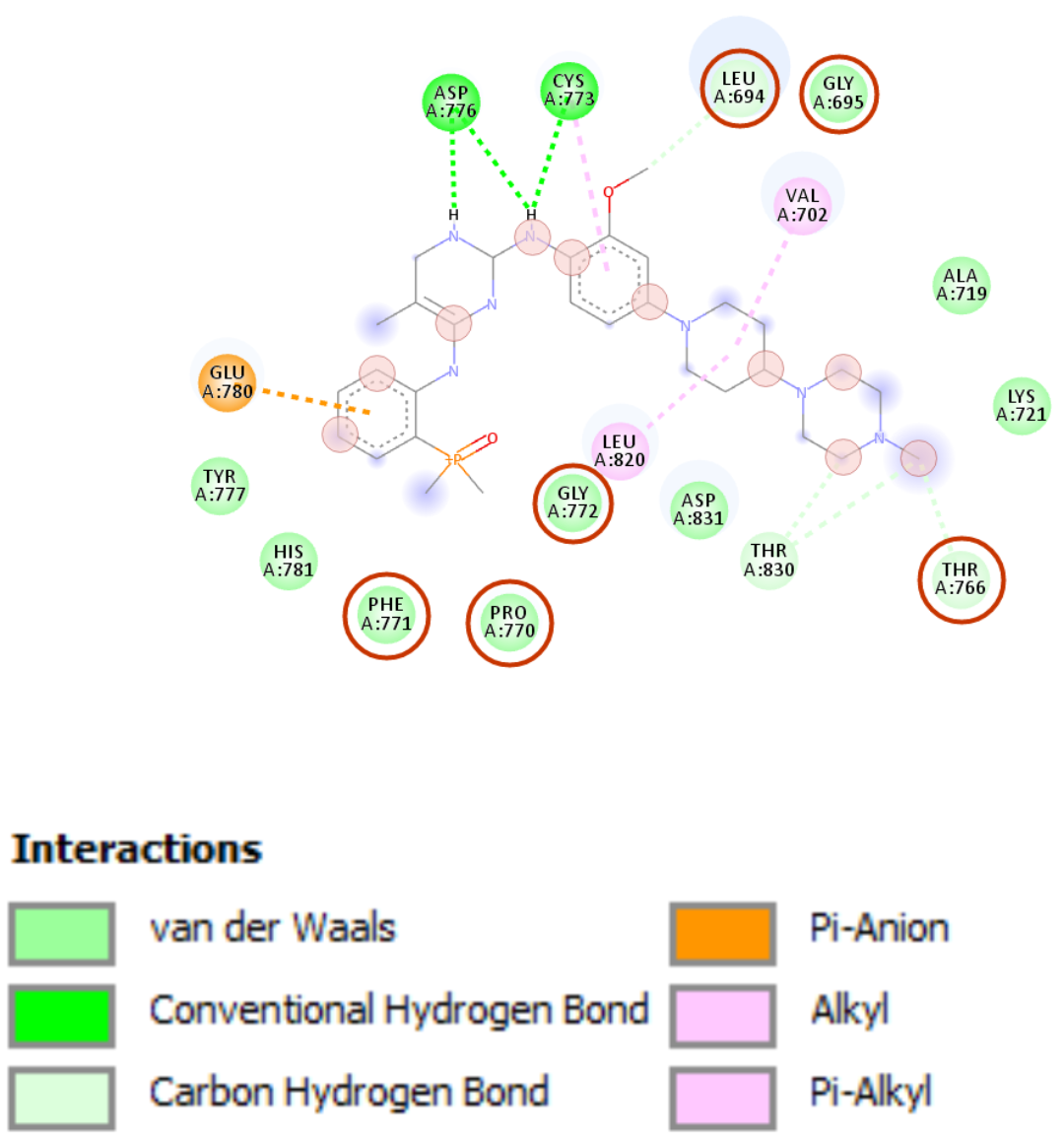}
	}\hspace{1em}%
	\\
	\subcaptionbox{\label{fig:zanubrutinib}}
	{
		\includegraphics[scale=.2]{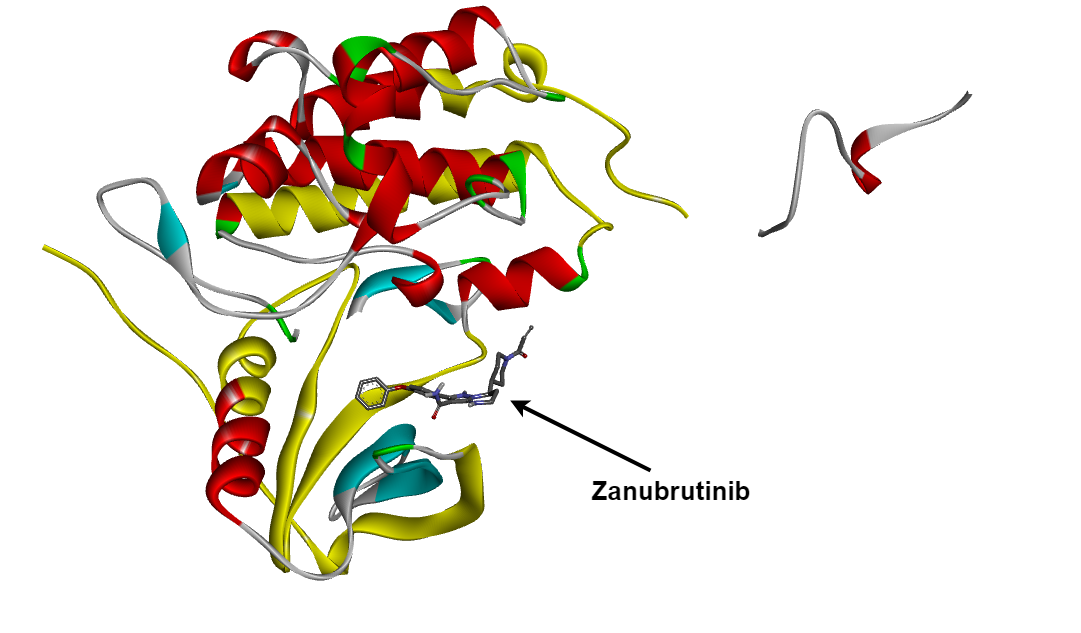}
		\includegraphics[scale=.2]{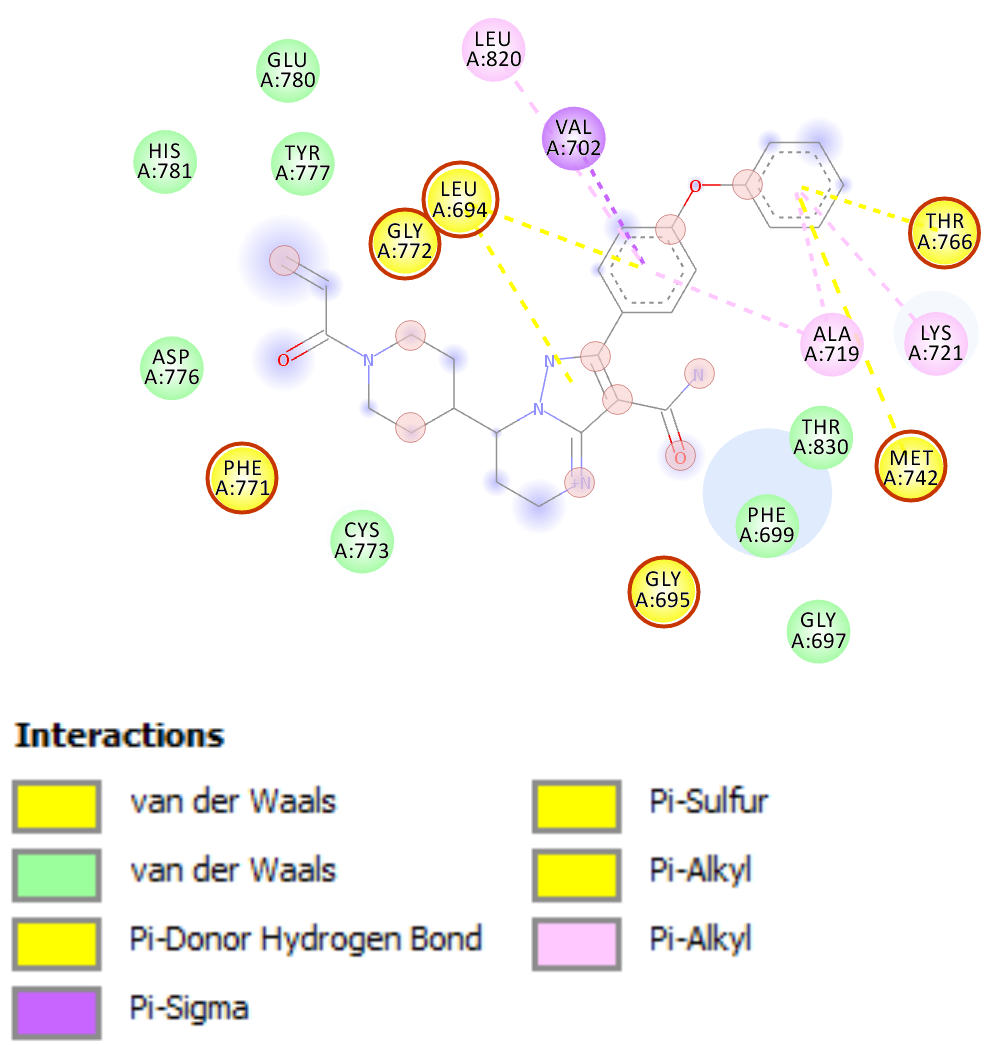}
	}\hspace{1em}%
%	\\
%	\subcaptionbox{\label{fig:cpi_interp}}
%	{
%	\includegraphics[scale=.2]{CPI_EGFR_1M17_Brigatinib_DB12267_complex_cpi}
%	}\hspace{1em}%

	%	}
	
	\caption{Epidermal Growth Factor Receptor (EGFR-1M17) tyrosine kinase domain in complex with (a) Brigatinib and (b) Zanubrutinib. The amino acid residues in yellow represent the top-10 subsequences predicted by the JoVA1 model. For both complexes, the corresponding interaction analysis of the ligand in the binding pocket of EGFR-1M17 is shown on the right. The top-10 atoms of ligand predicted by the JoVA1 model to be influential in the interaction are depicted in transparent red circles. The amino acids shown in the interaction analysis and also among the top-10 residues in each complex are highlighted using red circles as borders.}
	\label{fig:interpretability_cs}
\end{figure}

\begin{figure}
	\centering
	\includegraphics[width=0.7\linewidth]{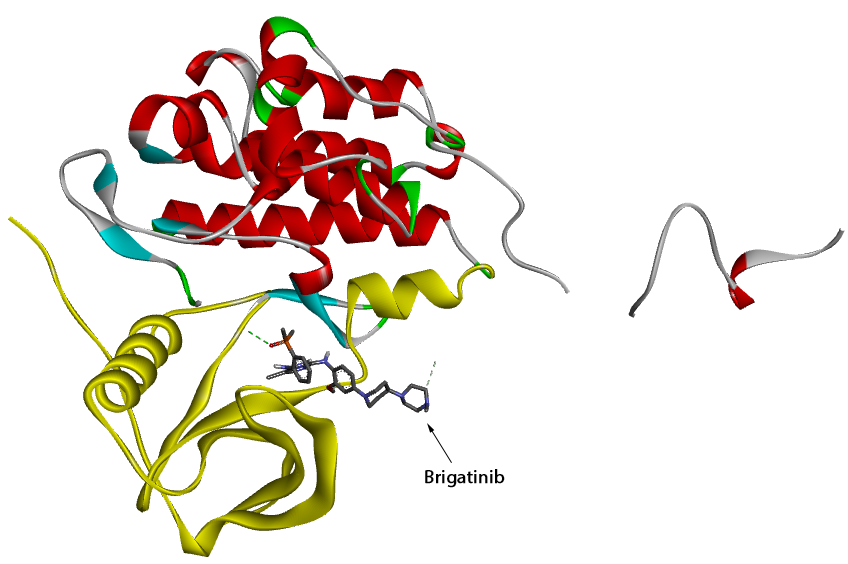}
	\caption{Mapping of CPI-Reg Protein Convolutional Neural Attention weights to 3D structure of EGFR-1M17 in complex with Brigatinib.}
	\label{fig:cpi_interp}
\end{figure}

In this section, we discuss a case study performed using the Drugbank~\cite{Wishart2018} database. The JoVA1 model trained on the KIBA dataset using the warm split scheme was selected to evaluate the ability of our approach to predict novel and existing interactions.

The human Epidermal Growth Factor Receptor (EGFR) was selected to be the target for the case study. While other targets could equally be chosen, EGFR was selected since it is implicated in breast cancer and is a popular target for cancer therapeutics. As regards this Drugbank case study, we refer to both the approved and investigational drug relations of EGFR as interactions.

We downloaded $13,339$ compounds from the Drugbank database containing $30$ interaction records for EGFR. Since the Drugbank database contains small and biological molecules, we filtered out all biologics. The filtered dataset contained $10,630$ small molecules, of which $21$ are known to target EGFR. Also, we removed all compounds that are present in the KIBA dataset to ensure that all drugs used for the case study were not part of the training set. As a result, the size of the final Drugbank dataset used for this case study was $9,484$, with $8$ EGFR interactions. Thus, $13$ of the $21$ small molecules in the Drugbank database are also present in the KIBA dataset. 

In table~\ref{tab:drugbank_cs} we present the top-50 predictions of the JoVA model. The model was able to predict. $6$ of the $8$ EGFR interactions in its first $50$ drugs, ranked according to the KIBA score. Also, it can be seen that the predicted KIBA scores for all the reported drugs fall under the KIBA value$\leq3.0$ threshold used in~\cite{Pahikkala2015} to indicate true interactions. Using the unfiltered $10, 630$ small molecules, the predicted KIBA values of $17$ of the $21$ EGFR interactions were all below the threshold mentioned above, with the remaining $4$ falling under $4.0$.

While these results demonstrate the ability of our proposed approach to improve the virtual screening stage of drug discovery, the novel predictions reported herein could become possible cancer therapeutics upon further investigations.
\begin{table}
	\centering
	\caption{The top 50 drugs predicted to interact with the Epidermal Growth Factor Receptor by the JoVA1 model. Entries in \textbf{bold print} are drugs reported to target EGFR in the Drugbank database. The chemical formula of a drug is used if the name of the drug is long.}
	\label{tab:drugbank_cs}
	\resizebox*{!}{.93\textheight}{		
		\begin{tabular}{|l|l|l|l|}
			\hline
			\multicolumn{1}{|c|}{\textbf{Rank}} & \multicolumn{1}{c|}{\textbf{Drugbank ID}} & \multicolumn{1}{c|}{\textbf{Drug}} & \multicolumn{1}{c|}{\textbf{\begin{tabular}[c]{@{}c@{}}KIBA \\ score\end{tabular}}} \\ \hline
			1                                   & DB11963                          & \textbf{Dacomitinib}               & 1.314                                                                               \\ \hline
			2                                   & DB06021                          & \textbf{AV-412}                    & 1.516                                                                               \\ \hline
			3                                   & DB07788                          & $C_{19}H_{22}O_7$                           & 1.693                                                                               \\ \hline
			4                                   & DB12818                          & NM-3                               & 1.775                                                                               \\ \hline
			5                                   & DB14944                          & Tarloxotinib                       & 1.834                                                                               \\ \hline
			6                                   & DB02848                          & $C_{22}H_{22}N_{4}O_{3}S$                        & 1.901                                                                               \\ \hline
			7                                   & DB05944                          & \textbf{Varlitinib}                & 1.912                                                                               \\ \hline
			8                                   & DB12669                          & 4SC-203                            & 1.915                                                                               \\ \hline
			9                                   & DB12114                          & Poziotinib                         & 1.993                                                                               \\ \hline
			10                                  & DB14993                          & Pyrotinib                          & 1.997                                                                               \\ \hline
			11                                  & DB06346                          & Fiboflapon                         & 2.172                                                                               \\ \hline
			12                                  & DB11652                          & Tucatinib                          & 2.301                                                                               \\ \hline
			13                                  & DB01933                          & 7-Hydroxystaurosporine             & 2.414                                                                               \\ \hline
			14                                  & DB12381                          & Merestinib                         & 2.423                                                                               \\ \hline
			15                                  & DB07270                          & $C_{20}H_{15}Cl_2N_3O_4S$                     & 2.467                                                                               \\ \hline
			16                                  & DB06469                          & Lestaurtinib                       & 2.534                                                                               \\ \hline
			17                                  & DB13517                          & Angiotensinamide                   & 2.582                                                                               \\ \hline
			18                                  & DB09027                          & Ledipasvir                         & 2.591                                                                               \\ \hline
			19                                  & DB11747                          & Barasertib                         & 2.645                                                                               \\ \hline
			20                                  & DB12668                          & Metenkefalin                       & 2.654                                                                               \\ \hline
			21                                  & DB03482                          & $C_{28}H_{36}N_{10}O_{15}P_2$                     & 2.692                                                                               \\ \hline
			22                                  & DB11613                          & Velpatasvir                        & 2.693                                                                               \\ \hline
			23                                  & DB07321                          & $C_{20}H_{15}Cl_2N_{3}O_5S$                     & 2.706                                                                               \\ \hline
			24                                  & DB03005                          & $C_{42}H_{45}N_8$                           & 2.708                                                                               \\ \hline
			25                                  & DB13088                          & AZD-0424                           & 2.708                                                                               \\ \hline
			26                                  & DB12673                          & ATX-914                            & 2.712                                                                               \\ \hline
			27                                  & DB12267                          & \textbf{Brigatinib}                & 2.717                                                                               \\ \hline
			28                                  & DB11973                          & Tesevatinib                        & 2.721                                                                               \\ \hline
			29                                  & DB12706                          & Seletalisib                        & 2.724                                                                               \\ \hline
			30                                  & DB15343                          & HM-43239                           & 2.755                                                                               \\ \hline
			31                                  & DB12183                          & Sapitinib                          & 2.764                                                                               \\ \hline
			32                                  & DB15035                          & \textbf{Zanubrutinib}              & 2.764                                                                               \\ \hline
			33                                  & DB15168                          & Cilofexor                          & 2.772                                                                               \\ \hline
			34                                  & DB06915                          & $C_{10}H_{8}O_5$                            & 2.777                                                                               \\ \hline
			35                                  & DB11853                          & Relugolix                          & 2.778                                                                               \\ \hline
			36                                  & DB15407                          & Acalisib                           & 2.797                                                                               \\ \hline
			37                                  & DB05038                          & Anatibant                          & 2.821                                                                               \\ \hline
			38                                  & DB14795                          & AZD-3759                           & 2.837                                                                               \\ \hline
			39                                  & DB06638                          & Quarfloxin                         & 2.837                                                                               \\ \hline
			40                                  & DB01763                          & $C_{21}H_{28}N_{7}O_{16}P_3S$                     & 2.857                                                                               \\ \hline
			41                                  & DB15403                          & Ziritaxestat                       & 2.859                                                                               \\ \hline
			42                                  & DB12557                          & FK-614                             & 2.859                                                                               \\ \hline
			43                                  & DB07838                          & $C_{17}H_{12}N_2O_4S_2$                       & 2.864                                                                               \\ \hline
			44                                  & DB11764                          & Spebrutinib                        & 2.866                                                                               \\ \hline
			45                                  & DB07698                          & $C_{18}H_{14}ClN_5$                         & 2.869                                                                               \\ \hline
			46                                  & DB13164                          & \textbf{Olmutinib}                 & 2.879                                                                               \\ \hline
			47                                  & DB12064                          & BMS-777607                         & 2.912                                                                               \\ \hline
			48                                  & DB09183                          & Dasabuvir                          & 2.914                                                                               \\ \hline
			49                                  & DB06666                          & Lixivaptan                         & 2.934                                                                               \\ \hline
			50                                  & DB06734                          & Bafilomycin B1                     & 2.937                                                                               \\ \hline
		\end{tabular}
	}
\end{table}

\subsection{Interpretability Case Study}\label{sec:interpretability}
The interpretability of DTI predictions is important to the drug discovery process. The ability to interpret an interaction in both the compound and target directions of the complex could reveal abstract intermolecular relationships.

Therefore, we performed an interpretability case study using Brigatinib and Zanubrutinib as the ligands and EGFR (Protein Data Bank ID: 1M17) as the macromolecule in two case studies. The EGFR structure was retrieved from the PDB\footnote{https://www.rcsb.org/structure/1M17} and the ligand structures from the DrugBank for docking experiments. We used PyRx~\cite{Dallakyan2015} to perform in-silico docking and Discovery Studio (v20.1.0) to analyze the docking results. We then mapped the top-10 atoms and top-10 amino acid residues predicted by the JoVA1 model used in the Drugbank case study above unto the docking results. The attention outputs of the model were used in selecting these top-k segments. In Figure~\ref{fig:interpretability_cs}, the yellow sections of the macromolecule indicate the top-10 amino acid residues, whereas the top-10 atoms of the ligand are shown in red transparent circles in the interaction analysis results on the right of each complex.

In the case of the EGFR-Brigatinib complex (see Figure~\ref{fig:brigatinib}), we realized that the selected amino acid residues were mostly around the binding pocket of the complex. While we show only the best pose of the ligand in Figure~\ref{fig:interpretability_cs}, the other selected amino acid residues were identified by the docking results to be for other poses of the ligand. Also, selected atoms of the ligand happen to be either involved in an intermolecular bond or around regions identified by the docking results analysis to be essential for the interaction. Interestingly, the amino acids of the macromolecule identified to be intimately involved in the interaction and also among the top-10 residues are predominantly in a Van der Waals interaction with the ligand. Thus, the model considered stability of the interaction at the active site to be significant in determining the binding affinity.

Likewise, the EGFR-Zanubrutinib case study yielded interpretable results upon examination. It could be seen in Figure~\ref{fig:zanubrutinib} that the top-10 amino acid residues selected in the EGFR-Brigatinib case study were identified again. Thus, the model has learned to consistently detect the binding site in both case studies. Indeed, this consistency was also observed in several other experiments using EGFR-1M17 and other ligands\footnote{https://git.io/JJDVC}. This aligns with knowledge in the domain where an active site could be targeted by multiple ligands. The highlighted top-10 amino acid residues also contain three phosphorylation sites (Thr686, Tyr740, Ser744), according to the NetPhos 3.1~\cite{Blom1999}\footnote{https://services.healthtech.dtu.dk/} server prediction results. 

Additionally, the interaction analysis of the EGFR-Zanubrutinib case study reveals that a number of the amino acids selected in the top-10 segments are involved in pi-interactions which are vital to protein-ligand recognition. We also note that some of the selected atoms of Zanubrutinib are in the aromatic regions where these pi-interactions take place. In another vein, other selected amino acids are involved in Van der Waals interactions which reinforce the notion of stability being significant in determining the binding affinity.

In another vein, we mapped the attention weights of a CPI-Reg model trained on the KIBA dataset onto the subsequences of EGFR. The EGFR target was paired with each of the DrugBank compounds and served as input to the CPI-Reg model. In Figure~\ref{fig:cpi_interp} we show the complex of EGFR and Brigatinib with the top-10 subsequences, as defined by the attention weights, highlighted in yellow. It should be noted that, unlike JoVA, the CPI-Reg attention mechanism enables interpretability only for the target. While the highlighted subsequence in Figure~\ref{fig:cpi_interp} seems to show one of the binding sites detected by JoVA (see Figure~\ref{fig:interpretability_cs}), we realized upon analysis of the CPI-Reg model that equal weights were assigned to all subsequences of EGFR. Therefore, the top-10 could be any 10 subsequences of EGFR and hence not helpful for interpretability analysis. We observed this phenomenon for all other DrugBank compounds\footnote{\url{https://git.io/JJAqZ}}. Further analysis revealed that poor representation of EGFR was largely the reason for this undesirable behavior. This is not surprising since CPI-Reg performed poorly across the RMSE, CI, and $R^2$ metrics as evaluated on the benchmark datasets.

In the nutshell, our approach is also able to offer biologically plausible cues to experts for understanding DTI interactions. Such an ability could be invaluable in improving existing virtual screening methods in rational drug discovery.

\section{Conclusion}\label{sec:conclusion}
In this study, we have discussed the significance of studying DTI as a regression problem and also highlighted the advantages that lie within leveraging multiple entity representations for DTI prediction. Our experimental results indicate the effectiveness of our proposed self-attention based method in predicting binding affinities and offering biologically plausible interpretations via the examination of the attention outputs. The ability to learn rich representations using the self-attention method could have applications in other cheminformatic and bioinformatic domains such as drug-drug and protein-protein studies.  

\section*{Acknowledgement}
We would like to thank Siqing Zhang and Chenquan Huang for their help in setting up the experiment platforms. We are also grateful to Orlando Ding, Obed Tettey Nartey, Daniel Addo, and Sandro Amofa for their insightful comments. We thank all reviewers of this study.

\section*{Funding}
This work was partly supported by SipingSoft Ltd., China.

\clearpage
\bibliography{biblio}

\end{document}